# Transfusor: Transformer Diffusor for Controllable Human-like Generation of Vehicle Lane Changing Trajectories

**Jiqian Dong**
Graduate Research Assistant, Center for Connected and Automated Transportation (CCAT), and Lyles School of Civil Engineering, Purdue University, West Lafayette, IN, 47907.
Email: dong282@purdue.edu
ORCID #: 0000-0002-2924-5728

**Sikai Chen\***
Assistant Professor, Department of Civil and Environmental Engineering, University of Wisconsin-Madison, Madison, WI, 53706.
Email: sikai.chen@wisc.edu
ORCID #: 0000-0002-5931-5619
(Corresponding author)

**Samuel Labi**
Professor, Center for Connected and Automated Transportation (CCAT), and Lyles School of Civil Engineering, Purdue University, West Lafayette, IN, 47907.
Email: labi@purdue.edu
ORCID #: 0000-0001-9830-2071

## ABSTRACT

With ongoing development of autonomous driving systems and increasing desire for deployment, researchers continue to seek reliable approaches for ADS systems. The virtual simulation test (VST) has become a prominent approach for testing autonomous driving systems (ADS) and advanced driver assistance systems (ADAS) due to its advantages of fast execution, low cost, and high repeatability. However, the success of these simulation-based experiments heavily relies on the realism of the testing scenarios. It is needed to create more flexible and high-fidelity testing scenarios in VST in order to increase the safety and reliabilityof ADS and ADAS.To address this challenge, this paper introduces the "Transfusor" model, which leverages the transformer and diffusor models (two cutting-edge deep learning generative technologies). The primary objective of the Transfusor model is to generate highly realistic and controllable human-like lane-changing trajectories in highway scenarios. Extensive experiments were carried out, and the results demonstrate that the proposed model effectively learns the spatiotemporal characteristics of humans' lane-changing behaviors and successfully generates trajectories that closely mimic real-world human driving. As such, the proposed model can play a critical role of creating more flexible and high-fidelity testing scenarios in the VST, ultimately leading to safer and more reliable ADS and ADAS.

## INTRODUCTION

The autonomous vehicle (AV) is considered a crucial element of the next generation of transportation due to its potential to bring about several significant benefits. These prospective advantages include the reduction of human labor and associated costs, an improvement in safety and reliability on the roads, as well as a decrease in emissions and energy consumption (*1–8*). However, before reaching the stage of full automation, where AVs dominate the traffic stream (*5, 9–11*), there will be an extended period known as the "transition period." During this phase, the roadways will be shared by both autonomous vehicles (AVs) and human-driven vehicles (HDVs). This scenario is commonly referred to as mixed or heterogeneous traffic (*4, 12, 13*). Given this mixed traffic environment, it becomes imperative that all autonomous driving features, including autonomous driving systems (ADS) and advanced driver assistance systems (ADAS), undergo comprehensive testing. The purpose of such testing is to ensure the safety and reliability of these technologies under real-world conditions, where interactions between AVs and HDVs occur. Only after rigorous testing and validation can these autonomous driving features be confidently released onto the market. This diligent approach is essential to build trust among the public and stakeholders and to pave the way for a safer and more efficient transportation future (*14*).

### Virtual Simulation Test

Of the various approaches for testing ADS, the virtual simulation test (VST) stands out as the predominant method, primarily due to its cost-effectiveness in terms of human labor and cost, as well as its high repeatability. However, the full potential of VST is somewhat constrained by the level of realism and fidelity achievable in testing scenarios. In order to truly immerse the ego vehicle within realistic human driving scenarios, the background traffic must be carefully defined to capture the dynamics and diversity of human drivers. Unfortunately, bridging the gap between the simulated environment and the real world remains a challenge (*15–18*). For example, there always exists a well-known dilemma called the "curse of sparsity" (*17*), where certain driving behaviors, particularly safety-critical ones, are exceptionally rare. Therefore, incorporating these infrequent scenarios into the testing environment becomes difficult. Consequently, it may be the case that AV immersion in comprehensive non-safety-critical environments and passing all the experiments may still leave the AV inadequately prepared for hazardous road situations. This ultimately impedes the achievement of a comprehensive safety-oriented testing process.

### Vehicle Trajectory Generation

Vehicle trajectory generation involves the task of creating viable paths, predominantly utilized in the path planning module for autonomous vehicles. In the context of the virtual simulation testing (VST), this module is responsible for generating not only feasible but also human-like trajectories for other simulated agents (background traffic) to follow, thereby creating realistic testing scenarios. Consequently, the trajectory generation process in this work centers around two key requirements: human-likeness and controllability. Human-likeness entails the trajectory generator learning from the behavior of real human drivers, encompassing diverse driving styles, such as aggressiveness or conservatism, as well as capturing the distinctive characteristics of various vehicle types, including trucks and cars. Controllability emphasizes the generator's capacity to produce trajectories with desired properties. This flexibility ensures that the testing scenarios can be tailored and customized accordingly to suit specific needs and objectives.

Trajectory generation and trajectory prediction are closely related tasks, with the former being a generative model that directly models the data distribution, while the latter is a discriminative model focused on predicting data labels. However, the modeling approaches for both tasks can be shared (i.e., through deep learning models or traditional methods). Traditionally, trajectory prediction and generation have relied on classic statistical machine learning models like Gaussian Processes, hidden Markov models, and Bayesian Networks, as well as heuristics such as constant velocity, maneuver-based prediction, or other physics-based models. The two review papers (*19, 20*), summarize these methods.

In recent years, advanced AI technologies, particularly deep learning, have revolutionized trajectory prediction and generation due to their exceptional generalization capabilities. As reviewed in the literature (*20–22*), deep learning-based approaches have predominantly been applied as discriminative models. However, there are limited examples where they were used as generative models. Notably, since the time they were initially proposed in the literature (*23*), diffusion models have shown remarkable performance in various domains including image generation (*23, 24*), audio and speech generation (*25, 26*), robotics action generation (*27, 28*). Within the trajectory generation domain, a few recent works (*16, 29*) have successfully applied diffusion models in a context-aware manner to generate pedestrian trajectories and controllable vehicle trajectories.

Although these methodologies have achieved success, they were not explicitly tailored to simulate human-like lane-changing behaviors in highway scenarios, which are widely acknowledged as the primary contributors to collisions (*30*). Additionally, these pioneering approaches have overlooked the heterogeneity in driving styles among human drivers, leading to an inadequate representation of rich and comprehensive testing environments. Therefore, there is still room for further research and development in this area to address the modeling of human lane-changing behaviors more effectively.

**Main Contribution**

To bridge this gap, this paper combines the diffusor model and the state-of-the-art (SOTA) sequence-to-sequence (CTC) deep learning architecture, Transformer (*31, 32*), to yield the "Transfusor" model. The objective of the model is to generate lane-changing trajectories on the highway scenario for VST. The model is trained and evaluated on the **High**way **D**rone (HighD) dataset (*33*) that has over 7500 lane changing trajectories. Through the training processs, it is observed that the model can successfully learn characteristics of the labeled human-driven trajectories and can generate human-like unseen trajectories.

The main contribution of the work can be summarized as follows:

- Proposed a new deep learning architecture with a novel diffusion model (termed "Transfusor") using the Transformer model to generate controllable human-like lane-changing trajectories.
- Cleaned and analyzed the lane-changing trajectoreis of human drivers in the HighD dataset, and labeled them with the desired properties.
- Developed a noval evaluation metric to evaluate the generalizability and extendability of generative model.

**METHODS**

**Figure 1** illustrates the overall working process of the proposed Transfusor model. To ensure controllability, the model adopts a conditional trajectory diffusor structure, incorporating condition encoding to facilitate learning the association between each trajectory and its corresponding label. The condition encoding plays a crucial role during the generation process, guiding the diffusion mechanism to produce desirable trajectories. In this study, the input trajectories are tagged with specific labels denoting vehicle type (car or truck), lane changing direction (left or right), and aggressiveness level (low, normal, or high). These labels serve as essential guidance for the model, enabling it to generate trajectories that conform to the specified conditions.

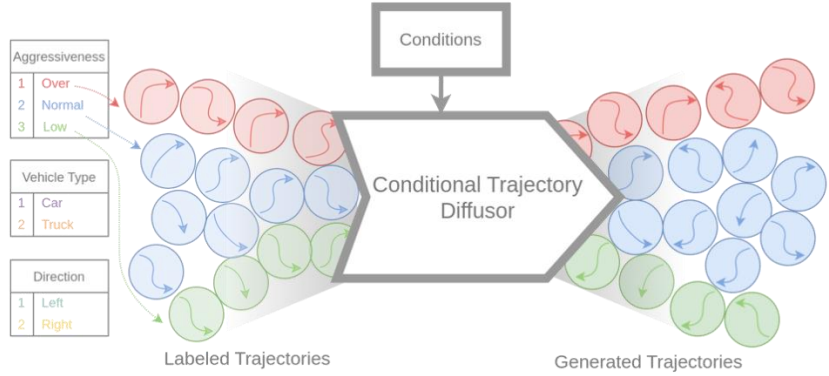

**Figure 1. Conditional Trajectory Diffusion Framework**

**Denoising Diffusion Probabilistic Models**

Inspired by concepts from non-equilibrium thermodynamics, Denoising Diffusion Probabilistic Models (DDPM), commonly referred to as diffusion models, belong to the category of deep generative models. The fundamental goal of these models is to learn the dynamics of a parameterized Markov chain, which gradually generates a target data distribution starting from a shared random distribution. This procedure centers around denoising, where noise is progressively removed through iterative steps. The diffusion process consists of both a forward process and a backward process.

*Forward Process*

With regard to the forward path, it takes a normal trajectory as the input and performs K steps of noise addition. More specifically, at each time step the forward process corrupts the trajectory by multiplying a new Gaussian distribution to the data obtained at previous step. Denoting the data distribution (the original trajectories in the dataset) $\mathbf{x}_0$, where $\mathbf{x}_0 \in \mathbb{R}^{T \times 2}$ and $T$ is the trajectory length, 2 represents the 2D location in the $(x, y)$, each step adds:

$$q(\mathbf{x}_k \mid \mathbf{x}_{k-1}) = \mathcal{N}\big(x_k; \sqrt{1 - \beta_k}\mathbf{x}_{k-1}, \beta_k \mathbf{I}\big) \qquad (1)$$

Therefore, the entire sequence can be represented as:

$$q(\mathbf{x}_{1:K} \mid \mathbf{x}_0) = \prod_{k=1}^{K} q(\mathbf{x}_k \mid \mathbf{x}_{k-1}) \qquad (2)$$

As all the noises follow Gaussian distribution, the diffusion process at any step k can be derived in a closed form as:

$$q(\mathbf{x}_k \mid \mathbf{x}_0) := \mathcal{N}\big(\mathbf{x}_k; \sqrt{\overline{\alpha_k}}\mathbf{x}_0, (1 - \overline{\alpha_k})\mathbf{I}\big) \qquad (3)$$

Where $\alpha_k = 1 - \beta_k$ and $\overline{\alpha_k} = \prod_{s=1}^{k} \alpha_s$ are the parameters after a reparameterization trick (*23*). As this forward path proceed (i.e., $K \to \infty$) the path is gradually corrupted and eventually approximating an isotropic non-informatic distribution, i.e., $\mathbf{x}_K \sim \mathcal{N}(\mathbf{0}, \mathbf{I})$.

*Backward Process*

The trajectory generation process is modeled through a reverse diffusion process from a non-informatic noise distribution. Again, this reverse process is characteristicod by parameterized Gaussian transitions. However, unlike the vanilla diffusion model (*23*) which takes only the previous step's trajectory as input, in this work, we add an additional condition indicating the trajectory class (i.e., truck, car, directions, aggressiveness). Here, we apply the same concept as the classifer-free guidance as proposed in (*34*) and incorporate the embeddings from the condition encoding $c$ as guidance as follows:

$$p_\theta(\mathbf{y}_{0:K} \mid \mathbf{c}) = p(\mathbf{x}_K) \prod_{k=1}^{K} p_\theta(\mathbf{x}_{k-1} \mid \mathbf{x}_k, \mathbf{c}) \qquad (4)$$

Where $p(\mathbf{x}_K)$ is the initial noise distribution, transition probability is:

$$p_\theta(\mathbf{x}_{k-1} \mid \mathbf{x}_k, \mathbf{c}) = \mathcal{N}\big(\mathbf{x}_{k-1}; \boldsymbol{\mu}_\theta(\mathbf{x}_k, k, \mathbf{c}); \boldsymbol{\Sigma}_\theta(\mathbf{x}_k, k, \boldsymbol{c})\big) \qquad (5)$$

The $\boldsymbol{\mu}_\theta(\mathbf{x}_k, k, \mathbf{c})$ and $\boldsymbol{\Sigma}_\theta(\mathbf{x}_k, k, \boldsymbol{c})$ are the outputs from the deep neural network parametrized by $\theta$. As derived in (*23*), $\boldsymbol{\Sigma}_\theta(\mathbf{x}_K, k, \boldsymbol{c}) = \sigma_k^2 \mathbf{I} = \beta_k \mathbf{I}$.

*Training Objective*

The objective for the diffusion process is to reconstruct the initial trajectories from the dataset. Therefore, the desired training goal is to maximize the log likelihood of $\mathbf{x}_0$, i.e., $\mathbb{E}[\log p_\theta(\mathbf{x}_0)]$. Following the derivation in (*23*), it is equivalent to maximize the variational lower bound, and the final loss function can be simplified to to compute the L2 distance between the known injected noise $\epsilon_k$ and predicted noise $\epsilon_{(\theta)}(\mathbf{x}_k, k, \mathbf{c})$ at each diffusion step:

$$L(\theta) = \mathbb{E}_{\epsilon, \mathbf{x}_0, k} \| \epsilon_k - \epsilon_{(\theta)}(\mathbf{x}_k, k, \mathbf{c}) \| \qquad (6)$$

**Multi-head Self-attention (MSA) Layer and Transformer Block**

The Transformer model incorporates a fundamental component called the multi-head self-attention (MSA) layer. MSA performs parallel computations of self-attention (SA), which employs dot product calculations to gauge the "similarity" between two inputs. Specifically, SA initially generates three representations known as query ($Q$), key ($K$), and value ($V$) using separate linear layers: $Q = XW_Q$; $K = XW_K$; and $V = X^T W_V$, where $W_Q$, $W_K$, $W_V$ represent the corresponding weights. Here, K and Q belong to the space $K, Q \in \mathbb{R}^{s \times d_k}$ and , $V \in \mathbb{R}^{s \times d_v}$, $d_k$ is the dimension for both keys and queries, and $d_v$ is the dimensions of and values, respectively. Subsequently, an attention score matrix ($a$) is created by taking the dot product of the query and the key, followed by applying softmax normalization. The attention score measures the correlation between different regions and is denoted as

$$a = softmax\left(\frac{QK^T}{\sqrt{d_k}}\right) \in \mathbb{R}^{s \times s} \qquad (7)$$

This attention mechanism allows the output embedding of each spatial location to contain not only the information from that specific location but also important details from other spatial locations. The attention scores serve as fusion weights for generating the attended feature maps. The output of the SA is obtained by taking elementwise dot product between the value ($V$) with the attention score ($a$) as follows:

$$head = Attention(K, Q, V) = a \odot V \qquad (8)$$

The MSA layer is essentially a parallel version of SA, wherein multiple SA operations are simultaneously computed by concatenating all the heads. This can be represented as

$$MSA(X) = concat(head_1, \dots, head_h)\, W_{out} \in \mathbb{R}^{s \times d_{out}} \quad (9)$$

where $W_{out} \in \mathbb{R}^{d_v \times d_{out}}$ denotes the weights for the final output linear layer, and h represents the number of heads. Compared to a single head, employing MSA enables each head to focus on different tasks simultaneously and attend to regions with varying scopes. This flexibility greatly enhances the model's generalization capabilities. The output from the MSA layer retains the same spatial dimension as the input feature map $X$. However, each spatial location contains "fused" information from itself as well as other regions, based on automatically computed correlations.

The Transformer block harnesses the MSA as its central layer, complemented by positional encoding to signify the sequential order of time series data, such as a trajectory. Furthermore, two layer-normalization layers and one feed-forward layer are seamlessly integrated within. To prevent the issue of gradient vanishing, a residual shortcut is included before the layer-normalization layer. These additional components collectively strengthen the Transformer block's capacity to effectively process and capture essential information from the input data. The architecture of the Transformer block is shown in **Figure 2. (a)**

### Condition Induced Linear Layer

As part of the primary motivation, each input trajectory sequence must include a "condition" that indicates the categories associated with the trajectory. The condition comprises two distinct features. Firstly, it encompasses the trajectory category, which is a combination of three sub-categories: lane change direction (left/right), vehicle type (truck/car), and aggressiveness level (less/normal/over). Altogether, there are twelve categories, and they are embedded in a high-dimensional space using an encoding layer. The second feature within the condition is the time embedding, which represents the time step of the diffusion model. This is accomplished through the utilization of a sinusoidal position encoding technique within the time embedding block. Subsequently, the category embedding, time embedding, and category embedding are concatenated, forming the condition embedding.

To merge the condition embedding with the trajectory features, a special linear layer called the condition-induced linear layer, inspired by (*29*) is employed. The condition-induced linear layer comprises three sets of weights ($W_1, W_2 \ and \ W_3$) and biases ($b_1 \ and \ b_2$). As shown in the **Figure XXX**. It takes the trajectory features ($x$) and condition embedding ($c$) as inputs. The overall computation logic of this layer can be summarized as follows:

$$f(\boldsymbol{x}, \boldsymbol{c}) = (W_1^T \boldsymbol{x} + \boldsymbol{b_1}) \odot \sigma(W_2^T \boldsymbol{x} + \boldsymbol{b_2}) + (W_3^T \boldsymbol{c} + \boldsymbol{b_3}) \qquad (10)$$

where $\sigma(\cdot)$ denotes the sigmoid function, which has the range of 0-1. Intuitively, the second sets of weight and bias ($W_2$ and $b_2$) represents a "gate" to control how much information is added to the output from the original linear layer ($W_1^T \boldsymbol{x} + \boldsymbol{b_1}$). Then an additional bias term is computed from the condition solely ($W_3^T \boldsymbol{c} + \boldsymbol{b_3}$). As such, this layer is a "weighted sum" of the information from both trajectory and condition, while the condition determines the fusion weights. This is the reason why this layer is refered as condition-induced linear layer.

**Overall Architecture**

The overall architecture of the Transfusor model is represented in **Figure 2(b)**. The model takes three raw inputs: trajectory, category, and time step, and processes them through several stages:

- Embedding: The trajectory, category, and time step inputs are first embedded into a higher-dimensional space using one linear layer and two encoding blocks. The category encoding involves transforming the categorical variable into numerical space using a default numerical embedding layer. The time encoding utilizes sinusoidal positional encoding, as previously employed in references (*31, 35*).

- Fusion: The three inputs (trajectory, category, and time step) are then fused together using a condition-induced linear layer. This process allows the model to incorporate the conditional information provided by the trajectory's category and time step.

- Transformer Block: The fused information is fed into a transformer block, which is responsible for learning and processing the data effectively, considering the interdependencies and contextual information.

- Dimension Reduction: After the transformer block, two additional dimension reduction condition-induced linear layers are applied to transfer the high-dimensional values back to the same dimension as the added noise in the forward process of the diffusion model.

- Loss Computation: The model computes the loss between the predicted noise and the known added noise from the forward process. For this research, a standard Mean Squared Error (MSE) loss function is utilized.

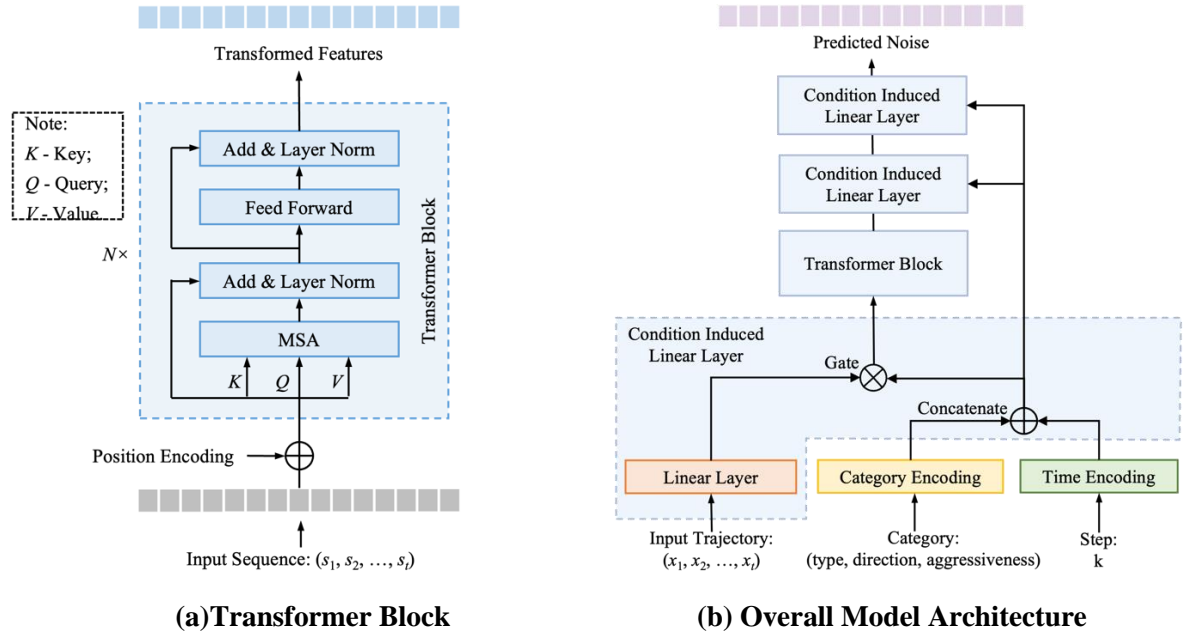

**(a) Transformer Block**          **(b) Overall Model Architecture**

**Figure 2. Model Structures**

## EXPERIMENT SETTINGS

### Dataset Preparation and Analysis

The HighD dataset (*33*) is a novel dataset comprising naturalistic human vehicle trajectories. It was captured by a drone and covers six distinct locations on German highways. The dataset encompasses trajectory data for over 110,000 vehicles and includes a total recording duration of 16.5 hours. A standout feature of the HighD dataset is its high framerate, capturing data at 25 Hz, which allows for a detailed examination of vehicle movements. Moreover, the dataset boasts impressive accuracy, with vehicle localization errors limited to less than 10 cm. In this particular study, the focus is on investigating lane-changing behaviors. To accomplish this, a data preprocessing step is specially devised. This preprocessing step involves extracting specific trajectory segments that pertain to lane-changing maneuvers. By narrowing the analysis to these segments, researchers can delve into the intricacies of lane-changing behavior in-depth, leveraging the richness of the HighD dataset to gain valuable insights.

#### *Lane Change Trajectory Extraction*

The extraction of lane changing trajectories from the overall vehicle trajectories in the HighD dataset involves two strategies: the fixed-window-size approach and the dynamic-window-size approach. For both methods, first, the crossing boundary time (CBT) is identified for each lane changing trajectory. The CBT is the exact timestep when the vehicle crosses the lane boundary, and it is determined by checking the "laneID" field in the trajectory data. If the "laneID" changes into the left or right lane, the CBT is recognized. Once the CBT is obtained, fixed-window-size approach assumes a fixed duration for each lane changing trajectory, consisting of two separate fixed time windows before and after the CBT. Two configurations were tested in the experiment: a) 3 seconds (75 frames) before and 3 seconds after CBT, resulting in a total of 6 seconds (150 frames). b) 6 seconds before and 6 seconds after CBT, resulting in a total of 12 seconds (300 frames). The advantage of this approach is its simplicity and computational efficiency. However, it oversimplifies the lane changing behaviors by assuming that all lane changes have the same duration, neglecting the heterogeneity of human driving behaviors.

In contrast to the fixed-window-size approach, the dynamic-window-size approach does not assume specific fixed time windows. This approach computes the starting time and ending time of each lane changing behavior based on the average lateral velocity ($v_y$) of the vehicle. The lane changing starting and ending timesteps are identified when the average lateral velocity over a moving time interval becomes stable (i.e., below the threshold of $0.2\ m/s$). In this work, we leverage the interval size of 1 second (25 frames). The dynamic-window-size approach captures the heterogeneity of human driving behaviors since the extracted lane changing trajectories may have different time durations, depending on the variability in lateral velocities.

Using both the fixed-window-size and dynamic-window-size approaches, the study ensures a comprehensive investigation of lane changing behaviors, considering both the simplicity of fixed time windows and the flexibility of capturing diverse behaviors with varying time durations. This approach allows for a more nuanced analysis of human lane-changing behaviors in the HighD dataset.

#### *Aggressiveness in Lane Changes*

Aggressive lane changes are characterized by rapid and abrupt movements between lanes, often at high speeds or with significant acceleration. These maneuvers may involve cutting off other vehicles without sufficient warning or space. In this work, we classify the aggressiveness of lane changing behaviors in the HighD dataset based on the ratio between mean lateral speed ($v_y$) and mean longitudinal speed ($v_x$) for each lane changing trajectory and label each trajectory with 3 aggressiveness level: low, normal, and over.

The classification strategy proceeds as follows: First, calculate the mean value ($\mu$) and standard deviation ($\sigma$) of the speed ratio $mean(v_y)/mean(v_x)$ for all the lane changing trajectories in the dataset.

Separate statistics are computed for cars and trucks. Second, the aggressiveness of each lane changing trajectory is classified based on the speed ratio, as follows:

- Over aggressive: If the speed ratio is greater than $\mu + \sigma$.

- Normal aggressive: If the speed ratio is between $\mu - \sigma$ and $\mu + \sigma$.

- Low aggressive: If the speed ratio is less than $\mu - \sigma$.

It is essential to consider separate statistics for cars and trucks because their lane changing behaviors differ. **Table 1**. provides the total statistics for the entire dataset using both lane changing extraction approaches, and the results show similar trajectory shapes and statistics. **Figure 3** presents the distributions of the speed ratio in the form of kernel density plot and box plots for different vehicle types and lane changing directions. Several observations can be made from these plots:

a) Trucks tend to have a higher speed ratio in lane changing compared to cars.

b) Truck right lane changes have higher variance than left lane changes and vehicles, indicating that trucks can exhibit very aggressive behaviors during some right lane changes, potentially due to certain imminent situations or roadblocks occurring on the rightmost lane.

c) There are many outliers in the car category for both left and right lane changes, suggesting the existence of very aggressive behaviors for cars. This could potentially increase the risk of accidents on the highway.

**Figure 4** visualizes all the lane changing trajectories, with the color indicating the aggressiveness category. For trajectories with higher aggressiveness, they tend to travel longer distances laterally while covering shorter distances longitudinally. These maneuvers may involve sharper turns with significant steering angles, increasing the risk of collisions, especially in highway scenarios.

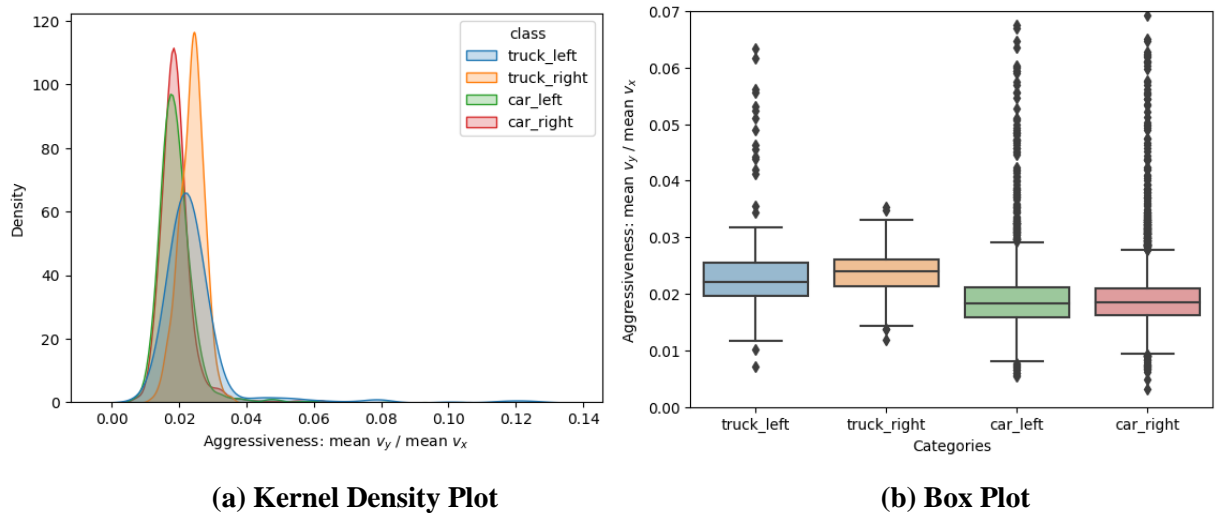

**(a) Kernel Density Plot**          **(b) Box Plot**

**Figure 3. Distribution of Aggressiveness of Lane Changing Trajectories**

**TABLE 1. Lane Changing Trajectory Statistics of HighD Dataset**

| Extracting methods | Directions | Veh. type | Avg. speed ratio | Std. speed ratio | # of trajs | # of trajs grouped by aggressivenes (less, normal, over) |
|---|---|---|---|---|---|---|
| | Left | Truck | 0.0248 | **0.0136** | 410 | (2, 384, 24) |
| Fixed window | | Car | 0.0193 | 0.0071 | 3051 | (130, 2733, 188) |
| | Right | Truck | 0.0237 | 0.0037 | 369 | (58, 260, 51) |
| | | Car | 0.0193 | 0.0063 | 3745 | (206, 3271, 268) |
| Dynamic window | Left | Truck | 0.0289 | **0.0159** | 366 | (2, 339, 25) |
| | | Car | 0.0237 | 0.0102 | 2600 | (78, 2337, 185) |
| | Right | Truck | 0.0270 | 0.0056 | 328 | (48, 233, 47) |
| | | Car | 0.0231 | 0.0086 | 2943 | (145, 2563, 235) |

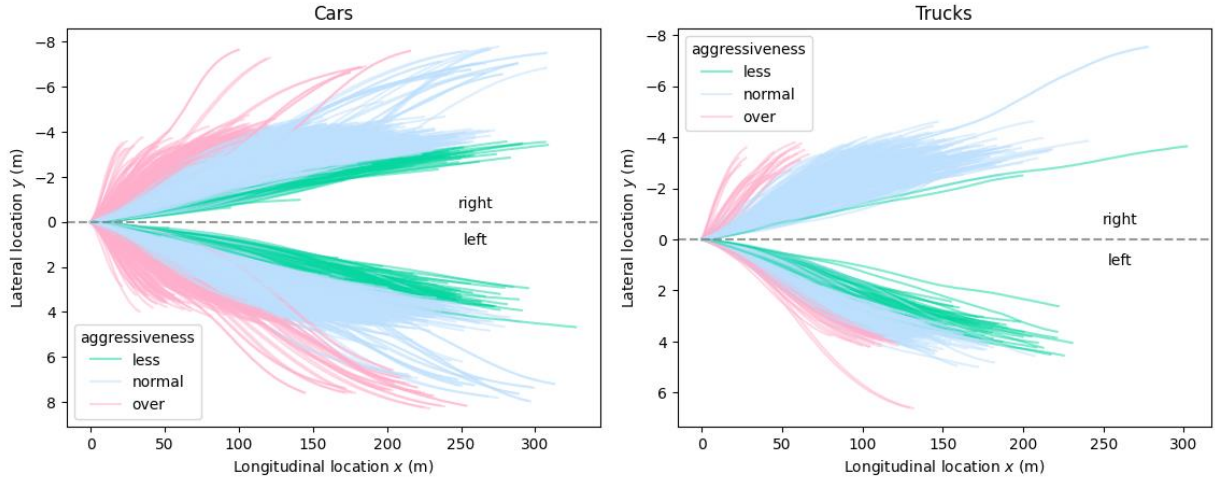

**(a) Car Lane Changes Trajectories**          **(b) Truck Lane Changes Trajectories**

**Figure 4. Lane Changing Trajectories for HighD Dataset**

**Baseline Model**

In this work, the baseline model utilized is the conditional variational autoencoder (CVAE) model. **Figure 5.** illustrates the structure of the baseline CVAE, which consists of a traditional encoder and decoder architecture. Similar to the diffusion model, the CVAE incorporates the same condition-induced linear layers to fuse the category features and two Transformer blocks individually for encoder and decoder. During training, the CVAE aims to minimize the reconstruction error between the input trajectories and the reconstructed trajectories. This is done by computing the mean squared error (MSE) between the two sets of trajectories. Additionally, the CVAE considers the KL divergence between the latent variables and a predefined uninformative distribution, which encourages the learned latent space to be more structured and interpretable. The total loss in the CVAE model is a weighted combination of the reconstructed error and the KL divergence error. The relative weights of these two components are usually set based on hyperparameter tuning and the specific requirements of the problem.

During the testing phase, the encoder is not used, and trajectories are generated solely by feeding the decoder with random samples from the latent variable space, combined with the encoded category information. This process allows the model to produce new and diverse trajectories based on the learned latent representations.

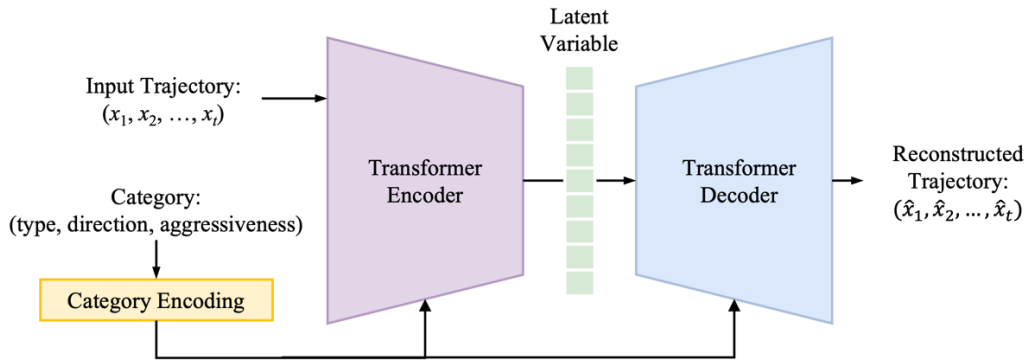

**Figure 5. Baseline Conditional Variational Autoencoder (CVAE) Model**

**Evaluation Metrics**

*Coverage Measure*

The Transfusor model is a generative model and not a predictive model. Therefore, in its evaluation, there are no ground-truth trajectories for direct comparison. As such, inspired by previous research (*36*), we propose the concept of coverage as the main evaluation metric. Coverage is a measure that reflects the degree of overlap between the generated trajectories from the model and the original trajectories in the dataset. **Figure 6** illustrates this concept. The total trajectory space, which represents the 2D drivable area. This space contains both the trajectories extracted from the dataset and the trajectories generated by the Transfusor model. The intersection between these two sets represents the "generated trajectories that resemble the dataset trajectories." To comprehensively evaluate the performance of the Transfusor model, two coverage values can be computed:

- Coverage ($c_1$): This value represents the proportion of the dataset trajectories that can be generated by the model. It is similar in concept to "recall" in binary prediction. A higher $c_1$ indicates that the model is successful in capturing and reproducing a large portion of the original dataset trajectories.

- Coverage ($c_2$): This value represents the proportion of the generated trajectories that can be found to have correspondence in the dataset. It is similar in concept to "precision" in binary prediction. A higher $c_2$ indicates that the generated trajectories have a higher likelihood of resembling actual trajectories from the dataset.

The ideal property for the generative model is to generate "human-like" but sometimes "unseen" trajectories, meaning that the generated trajectories should cover a large portion of the dataset while also having the ability to extend to some "unseen" scenarios. Therefore, $c_1$ and $c_2$ need to be balanced accordingly with the application context. Both coverage values ($c_1$ and $c_2$) are reported in the results section, using different criteria for determining the intersection between the generated and dataset trajectories.

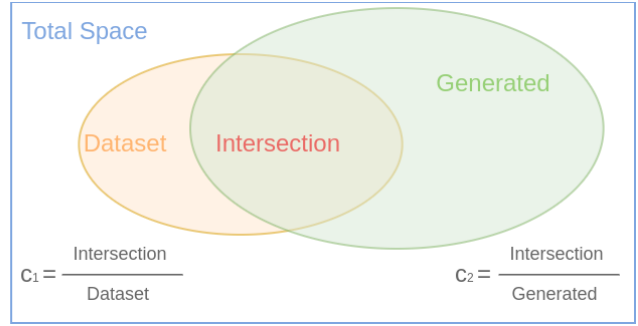

**Figure 5. Illustration of Coverage Computation**

*Trajectories Similarity*

In this work, to compute the intersection and find the correspondence between the trajectories in the original dataset and those generated by the Transfusor model, the average displacement error (ADE) is used as the similarity measure. The ADE is a standard error metric commonly used for trajectory prediction tasks. It quantifies the average L2 distance between all points in two trajectories as follows:

$$ADE = \frac{1}{T}\sum_{t=1}^{T} \sqrt{\left(\hat{x}_i^t - x_i^t\right)^2 + \left(\hat{y}_i^t - y_i^t\right)^2} \tag{11}$$

where $(\hat{x}_i^t, \hat{y}_i^t)$ and $(x_i^t, y_i^t)$ represent the predicted and original dataset trajectory points at timestep $t$, T is the overall trajectory length (in time steps). When the ADE falls below a certain threshold, it implies that the two trajectories share similar shape and characteristics. Consequently, these trajectories are considered as correspondences and are included in the "intersection" part of **Figure 5**.

*Implementation Details*

For training and evaluating our model, we utilize the trajectories obtained from the fixed-window size approach, specifically with a window size of 150 (equivalent to 6 seconds per lane changing). To enhance the training process, we downsample the dataset with a factor of 10, reducing the number of points in each trajectory to 15 (equivalent to a frame rate of 2.5Hz). To make the prediction more manageable and avoid scale-related issues in the output, we calculate the differences between each pair of points in the trajectories and only predict these incremental values.

Regarding the model architecture, we employ a 4-layer Transformer as the key neural network for the Transfusor model. The hidden dimension is set to 128, and each layer has 4 heads. The time encoding and condition encoding dimensions are both 64, and the encoding layer incorporates sinusoidal position encoding (*23*, *24*). Overall, the number of trainable parameters in the Transfusor model is 1,100,164 and the baseline CVAE model contains 1,918,084.

The model is trained for 2500 epochs over the entire dataset, comprising a total of 7575 lane changing trajectories. During training, we use a batch size of 128 and the Adam optimizer with an initial learning rate of 0.001. Remarkably, the model demonstrates efficient training on a single Nvidia GTX 1080 GPU, completing the process within 20 minutes. By employing these settings and configurations, we ensure the effective and scalable training of the Transfusor model, enabling it to learn and generate human-like lane-changing trajectories with good computational efficiency.

## RESULTS

### Comparative Results

From Table 2, we can observe the following insights regarding the coverage values for the CVAE model and the Transfusor model with ADE thresholds of 0.5 m and 1.0 m:

1. CVAE model: The CVAE model exhibits severe imbalance in the two coverage values, with $c_1$ (coverage of dataset trajectories generated by the model) being much smaller than $c_2$ (coverage of generated trajectories found in the dataset). This indicates that the CVAE model can only generate a small subset of the dataset trajectories. In Figure 5, the "intersection" segment representing trajectories that resemble the dataset only accounts for a small portion of the dataset, but it constitutes almost the entire space of the generated trajectories. Therefore, the CVAE model may struggle to generalize well to the diversity of lane-changing behaviors in the dataset.

2. Transfusor model: In contrast, the Transfusor model achieves a more balanced distribution between $c_1$ and $c_2$. This indicates that the Transfusor model has better performance in terms of generalizability and extendability compared to the CVAE model. The Transfusor model can generate trajectories that resemble a larger portion of the dataset while also having generated trajectories found within the dataset.

3. Rare cases: Both models face challenges in dealing with rare classes, which are classes with very few examples in the dataset. For example, the truck left low aggressiveness lane changes with only 2 trajectories in the dataset. The scarcity of training data in these rare classes can lead to difficulties for both models in capturing the characteristics of these specific lane-changing behaviors. As such, proper data augmentation techniques could be employed to mitigate this issue and improve the models' performance on rare classes.

In summary, the Transfusor model outperforms the CVAE model in terms of coverage values, demonstrating better generalization and performance. However, both models face challenges when dealing with rare classes due to limited training data, which could be addressed through data augmentation strategies.

**TABLE 2.**

| Veh. Type | Direction | Aggressiveness | Threshold ADE = 0.5 m | | | | Threshold ADE = 1.0 m | | | |
|---|---|---|---|---|---|---|---|---|---|---|
| | | | CVAE | | Diffusor | | CVAE | | Diffusor | |
| | | | $c_1$ | $c_2$ | $c_1$ | $c_2$ | $c_1$ | $c_2$ | $c_1$ | $c_2$ |
| car | left | less | 0.02 | 1.00 | 0.18 | 0.34 | 0.02 | 1.00 | 0.58 | 0.74 |
| | | normal | 0.01 | 1.00 | 0.61 | 0.94 | 0.05 | 1.00 | 0.71 | 0.98 |
| | | over | 0.01 | 0.94 | 0.06 | 0.50 | 0.03 | 1.00 | 0.41 | 0.74 |
| | right | less | 0.01 | 1.00 | 0.28 | 0.68 | 0.07 | 1.00 | 0.73 | 0.90 |
| | | normal | 0.02 | 1.00 | 0.27 | 0.96 | 0.07 | 1.00 | 0.72 | 0.98 |
| | | over | 0.00 | 0.00 | 0.14 | 0.40 | 0.01 | 1.00 | 0.19 | 0.52 |
| truck | left | less | 0.00 | 0.00 | 0.50 | 0.16 | 0.00 | 0.00 | 1.00 | 0.12 |
| | | normal | 0.07 | 1.00 | 0.73 | 0.78 | 0.21 | 1.00 | 0.88 | 0.98 |

| | | | | | | | | | | | |
|---|---|---|---|---|---|---|---|---|---|---|---|
| | | over | 0.00 | 0.00 | 0.25 | 0.54 | 0.08 | 1.00 | 0.58 | 0.62 |
| | right | less | 0.02 | 1.00 | 0.29 | 0.26 | 0.03 | 1.00 | 0.69 | 0.82 |
| | | normal | 0.10 | 1.00 | 0.79 | 0.88 | 0.22 | 1.00 | 0.93 | 1.00 |
| | | over | 0.10 | 1.00 | 0.63 | 0.62 | 0.25 | 1.00 | 0.90 | 0.84 |

## Qualitative Results

The visualizations of the generated trajectories from the Transfusor model and CVAE model are presented in Figure 6 and Figure 7. In these figures, 20 trajectories are generated for each of the 12 categories, and the trajectories are color-coded accordingly. In Figure 6, the trajectories generated by the Transfusor model exhibit highly diverse shapes and characteristics, accurately reflecting the heterogeneity of human drivers in different categories. The Transfusor model successfully learned the unique features and patterns of trajectories from each category, resulting in a wide variety of generated trajectories. On the other hand, Figure 7 shows the trajectories generated by the CVAE model. In contrast to the Transfusor model, the CVAE model produces trajectories that collapse into a small space, indicating that the model only learns a "mean" trajectory for each category. The lack of diversity in the generated trajectories from the CVAE model is consistent with the imbalanced coverage values observed in Table 2 and the conclusions drawn earlier.

The visualizations further support the previous findings that the Transfusor model greatly outperforms the CVAE model in capturing the heterogeneity of human lane-changing behaviors and generating more diverse and realistic trajectories. The Transfusor model's ability to learn distinct characteristics of each category leads to a broader range of generated trajectories, providing valuable insights into the human-like lane-changing behaviors in highway scenarios.

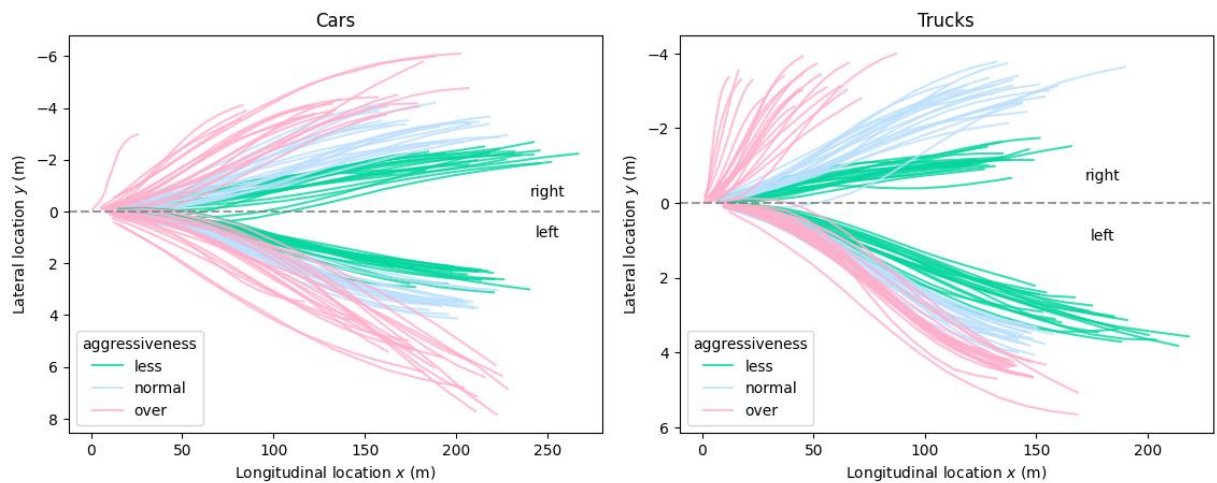

**Figure 6.   Trajectories Generated by Transfusor Model**

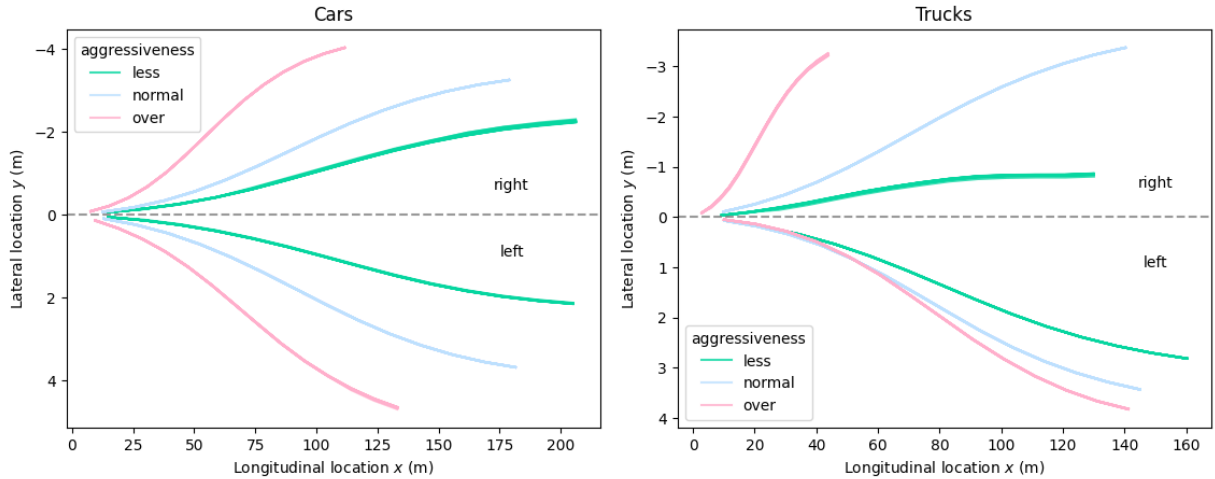

**Figure 7.  Trajectories Generated by Baseline CVAE Model**

**Visualization for Diffusion**

The diffusion model offers a significant advantage concerning model explainability. By visualizing the diffusion process, a crucial insight emerges: the drivable area progressively contracts from a completely random noise until only a confined space containing the desired trajectories remains. To illustrate this process, we generated contour plots, displayed in **Figure 8**, which depict the evolution. Each plot represents the distribution of 50 trajectories at a specific diffusion timestep $t$, computed using gaussian kernel density estimation. The color density in each plot represents the probability of the vehicle appearing at a specific location. Initially, the

As the diffusion progresses in the backward process from timestep 100 to 0, the noise is progressively filtered out, leading to a reduction in variance. Ultimately, this convergence results in the isolation of the desired trajectory. This characteristic can be leveraged in applications to generate highly dynamic trajectories during the initial stages of diffusion, effectively covering a larger drivable area. Consequently, it becomes possible to provide virtual simulation tests (VST) with a broader range of testing scenarios, enhancing the overall utility of the diffusion model.

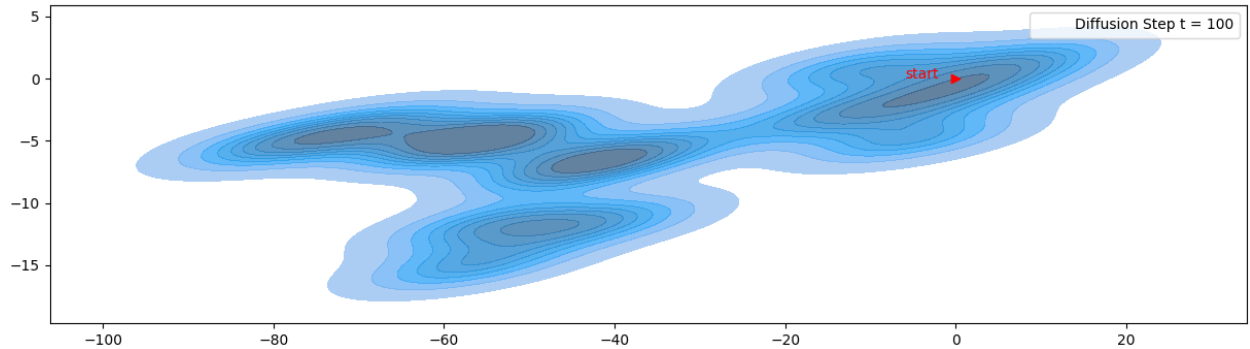

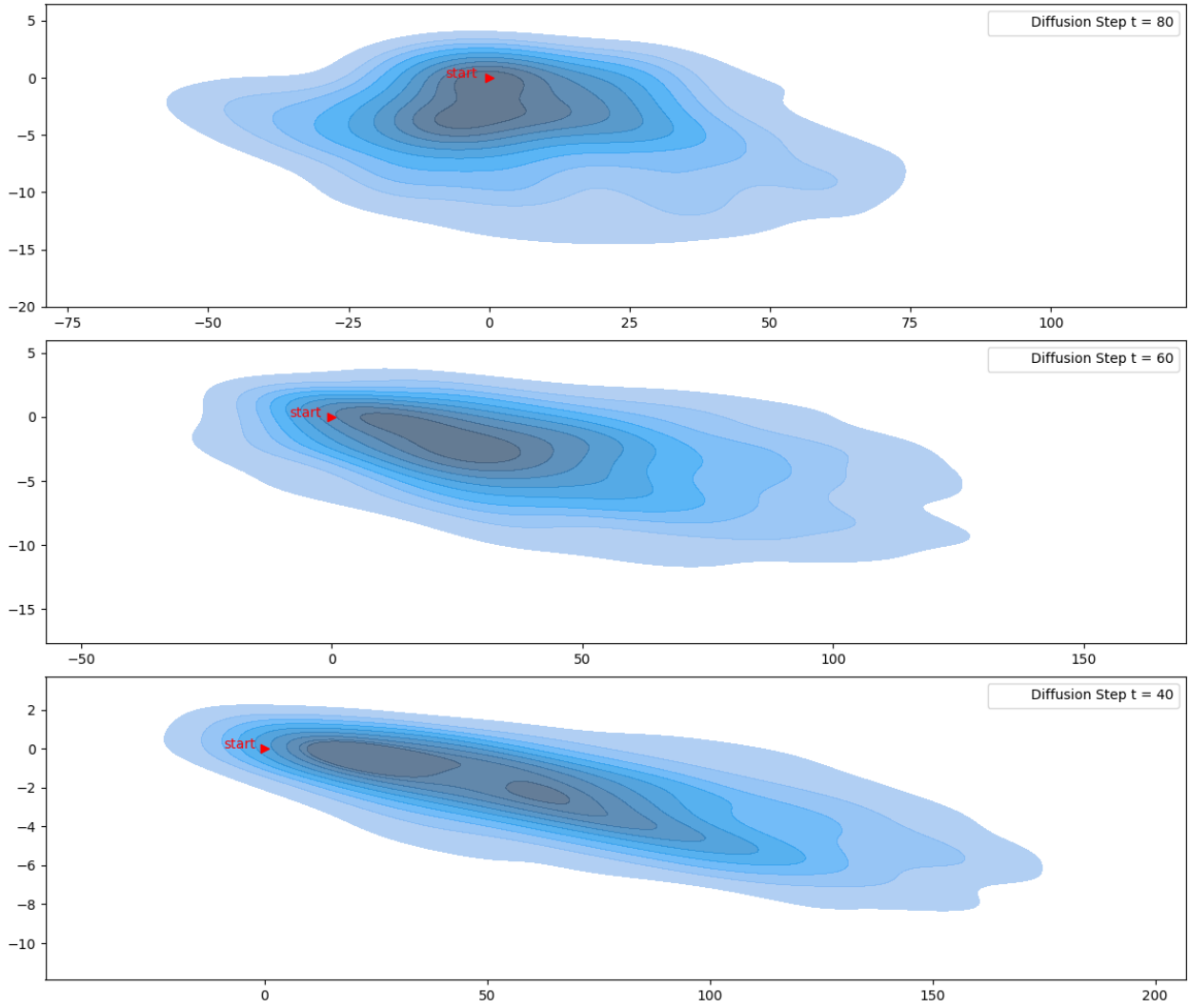

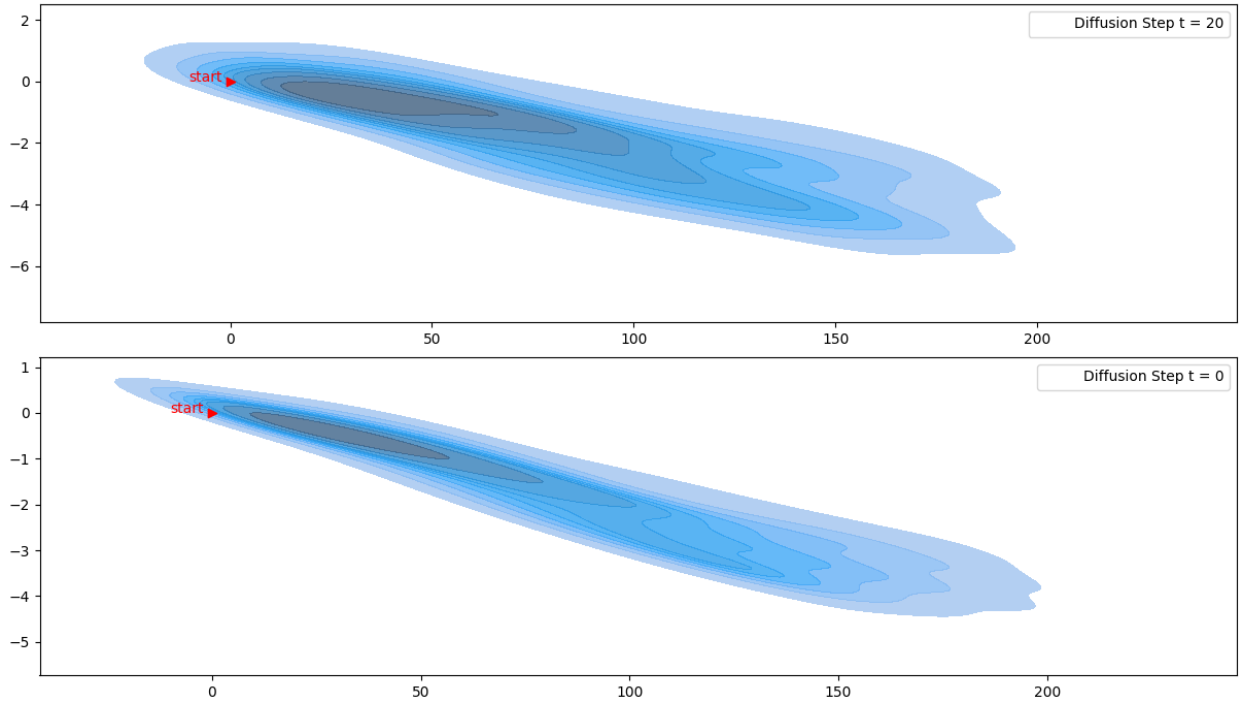

**Figure 8. Contour Plots for Diffusion Process (color intensity reflects the probability of the vehicle appearing at a specific location, darker represents high probability)**

## CONCLUSION

This paper introduces the Transfusor model, an innovative approach designed for generating human-like lane-changing trajectories in highway scenarios. Leveraging a conditional trajectory diffusor structure and condition encoding, our model achieves controllability in trajectory generation by effectively learning the correspondence between trajectories and their associated labels. In comparison to the baseline CVAE model, the Transfusor model exhibits superior performance, as evidenced by its balanced coverage values, signifying enhanced generalizability and extendability.

The Transfusor model's key strengths lie in its ability to produce human-like lane-changing behaviors in a controlled manner, generating trajectories with a diverse range of shapes. Moreover, the model's capacity to output intermediate diffusion results further enhances its utility as a potent generative tool for virtual simulation tests (VST) in autonomous driving research and traffic analysis.

In summary, the Transfusor model excels in generating realistic and varied lane-changing trajectories, effectively capturing the intricacies of human driving behaviors within highway scenarios. Future research endeavors may focus on refining the model's performance for rare classes through data augmentation techniques and exploring its applications in other real-world driving scenarios, opening up new avenues for advancements in autonomous driving and traffic management.

## ACKNOWLEDGMENTS

This work was supported by Purdue University's Center for Connected and Automated Transportation (CCAT), a part of the larger CCAT consortium, a USDOT Region 5 University Transportation Center funded by the U.S. Department of Transportation, Award #69A3551747105. The contents of this paper reflect the views of the authors, who are responsible for the facts and the accuracy of the data presented herein, and do not necessarily reflect the official views or policies of the sponsoring organization.

This manuscript is herein submitted for PRESENTATION ONLY at the 2024 Annual Meeting of the Transportation Research Board.

## AUTHOR CONTRIBUTIONS

The authors confirm contribution to the paper as follows: all authors contributed to all sections. All authors reviewed the results and approved the final version of the manuscript.